\documentclass[10pt, a4paper]{article}
\usepackage{booktabs}
\usepackage[hyphens]{url}
\usepackage{hyperref}
\hypersetup{colorlinks=true,breaklinks=true}
\usepackage{multirow}
\usepackage{color, colortbl}
\usepackage{xcolor}
\usepackage{inconsolata}
\usepackage{xspace,mfirstuc,tabulary}
\definecolor{Color}{gray}{0.9}
\usepackage{authblk}

\usepackage[colorinlistoftodos]{todonotes}
\newcommand*{\masakhanews}{\textbf{MasakhaNEWS} \xspace}

\usepackage{lrec-coling2024} 
\newcommand\blfootnote[1]{%
  \begingroup
  \renewcommand\thefootnote{}\footnote{#1}%
  \addtocounter{footnote}{-1}%
  \endgroup
}

\title{EthioLLM: Multilingual Large Language Models \\ for Ethiopian Languages
with Task Evaluation}

\name{\normalsize Atnafu Lambebo Tonja$^{1,2,\ast, \dagger}$, Israel Abebe Azime$^{3,\ast, \dagger}$, Tadesse Destaw Belay$^{1,\dagger}$,  \\
\textbf{\normalsize Mesay Gemeda Yigezu$^{1, \dagger}$, Moges Ahmed Mehamed$^{4, \dagger}$, Abinew Ali Ayele$^{5,6,\dagger}$,} \\
\textbf{\normalsize  Ebrahim Chekol Jibril$^{7,\dagger}$, Michael Melese Woldeyohannis$^{8,\dagger}$, Olga Kolesnikova$^{1}$, } \\
\textbf{\normalsize Philipp Slusallek$^{3}$, Dietrich Klakow$^{3}$,  Shengwu Xiong$^{4}$, }\\
\textbf{\normalsize Seid Muhie Yimam$^{6,\dagger}$} \\\\
\footnotesize
\address { $^\dagger$ Ethio NLP, $^1$ Instituto Politécnico Nacional, Mexico, $^2$ Lelapa AI, $^3$ Saarland University, Germany, 
\\
\footnotesize
$^4$  Wuhan University of Technology, China, $^5$ Bahir Dar University, Ethiopia, $^6$ Universität Hamburg, Germany, 
\\
 \footnotesize
 $^7$ Istanbul Technical University, Turkey, $^8$ Addis Ababa University, Ethiopia
}} 
\abstract{
Large language models (LLMs) have gained popularity recently due to their outstanding performance in various downstream Natural Language Processing (NLP) tasks. However, low-resource languages are still lagging behind current state-of-the-art (SOTA) developments in the field of NLP due to insufficient resources to train LLMs. Ethiopian languages exhibit remarkable linguistic diversity, encompassing a wide array of scripts, and are imbued with profound religious and cultural significance. This paper introduces EthioLLM -- multilingual large language models for five Ethiopian languages (Amharic, Ge'ez, Afan Oromo, Somali, and Tigrinya) and English, and Ethiobenchmark -- a new benchmark dataset for various downstream NLP tasks. We evaluate the performance of these models across five downstream NLP tasks. We open-source our multilingual language models, new benchmark datasets for various downstream tasks, and task-specific fine-tuned language models and discuss the performance of the models. Our dataset and models are available at the \href{https://huggingface.co/EthioNLP}{EthioNLP HuggingFace} repository.
 \\ \newline  }

\begin{document}

\maketitleabstract
\blfootnote{$^\ast$ Equal Contribution.}
\section{Introduction}
Large language models (LLMs) show a significant advancement in the field of artificial intelligence (AI) \cite{kasneci2023chatgpt}. In particular, the introduction of transformer \cite{vaswani2017attention} models has sparked the creation of powerful and effective multilingual pre-trained language models such as GPT \cite{brown2020language}, XLM-RoBERTa \cite{conneau2019unsupervised}, mT5 \cite{xue2020mt5}, and mBERT \cite{devlin2018bert}, which have attained cutting-edge performance in a variety of downstream NLP applications \cite{conneau2019unsupervised,devlin2018bert,alabi2022adapting,dossou2022afrolm,ogueji2021small,xue2020mt5}. These Pre-trained language models (PLMs) often outperform and may be tailored to a wide range of natural language processing (NLP) tasks \cite{kassner2021multilingual} including news classification \cite{adelani2023masakhanews}, machine translation \cite{wang2023document, lyu2023new}, sentiment analysis \cite{yadav2020sentiment, alsayat2022improving}, named entity recogination \cite{pan_etal_2017_cross}, part-of-speech tagging \cite{chiche2022part, nguyen2020phobert} and fake news detection \cite{kong2020fake, aggarwal2020classification}. However, a substantial portion of this development has been focused on high-resource languages. African languages have received especially less attention in this area  \cite{ogueji2021small}. Nevertheless, efforts are being made to address the challenges of low-resource languages, with a growing interest in developing Afro-centric models to improve NLP tasks for African languages. AfroLM \cite{dossou2022afrolm}, AfriBERTa \cite{ogueji2021small}, AmRoBERTa \cite{yimam2021introducing}, and AfroXLMR \cite{alabi2022adapting} aimed to bridge this gap by focusing on African languages, capturing their linguistic nuances, and improving language processing for these languages. However, those models have limitations as they did not cover most Ethiopian languages.
Ethiopia has over 85 spoken languages, but only a few have been included in developing NLP tasks and tools. Among these low-resource Ethiopian languages, there is a lack of pre-trained models and resources, which limits their ability to contribute to advancing AI research \cite{tonja2023natural,yimam2021introducing}.

In this paper, we introduce \textbf{EthioLLM} -- a multilingual pre-trained large language model for five Ethiopian languages with a new benchmark dataset for various downstream NLP tasks. 
Our contributions are as follows: \noindent \textbf{(1)} We introduce the first multilingual language models focusing on five Ethiopian languages and English. \noindent \textbf{(2)} We introduce \textbf{Ethiobenchmark} -- new benchmark datasets for various downstream NLP tasks. We compiled new datasets by amalgamating content from multiple sources to achieve broader language coverage for our study. Data sources for creating benchmark data and details about reconstruction are mentioned in Section \ref{benchmarkdataset}. All data releases were executed in consultation with the original authors if the data had not been released previously.
\noindent \textbf{(3)} We evaluate our models on existing datasets of MasakhaNEWS \cite{adelani2023masakhanews}, MasakhaNER \cite{adelani2021masakhaner}, AfriSenti \cite{muhammad2023AfriSenti,muhammadSemEval2023} and new benchmark datasets.
\noindent \textbf{(4)} We open-source\footnote{\url{https://github.com/EthioNLP/EthioLLM}} our multilingual language models, its training corpus, the new benchmark datasets, and the new task-specific fine-tuned models.  We aim to promote collaboration and streamline research and development for low-resource languages, especially within the context of Ethiopian languages. 

\section{Related Works}\label{related_works}

Some research efforts have been dedicated to creating multilingual language models that can be applied to low-resource languages, with the aim of mitigating the inequalities between languages with ample resources and those with limited resources.
Among prominent works, \citet{conneau2019unsupervised} introduced XLM-R, a multilingual masked language model trained on CommonCrawl\footnote{\url{https://commoncrawl.org/}} data for 100 languages, including three Ethiopian languages. \citet{feng2020language} presented a language-agnostic BERT sentence embedding (LaBSE) model supporting 109 languages, including three Ethiopian languages. \citet{devlin-etal-2019-bert} developed mBERT, a multilingual variant of
BERT trained in 104 languages, including four African languages. \citet{xue2020mt5} presented mT5, a massively multilingual pre-trained text-to-text transformer using a Common Crawl-based dataset covering 101 languages.

Geographic-based multilingual pre-trained language models have also been developed to address language under-representation \cite{ogueji2021small}. \citet{dossou2022afrolm} introduced AfroLM, a multilingual language model that employed a novel self-active learning framework entirely trained from scratch on a dataset encompassing 23 African languages, including two Ethiopian languages. \citet{ogueji2021small} presented AfriBERTa, a language model covering 11 African languages, including four Ethiopian languages.  \citet{alabi2022adapting} built Afro-XLMR by performing multilingual adaptive fine-tuning for 17 most-resourced African languages, including three Ethiopian languages and three other high-resource languages (Arabic, French, and English) widely spoken on the African continent to encourage cross-lingual transfer learning. Pre-training approaches for encoder-only models are extended to encoder-decoder models by introducing AfriTeVa, a pre-trained on 10 African languages from scratch \cite{jude-ogundepo-etal-2022-afriteva}. How to scale these encoder-decoder models to new languages and domains is investigated by \citet{adelani-etal-2022-thousand}, a multilingual language model covering 517 African languages.

For Ethiopian-centric languages, \citet{yimam2021introducing} introduced AmRoBERTa, a RoBERTa model trained using Amharic corpus. 

Most of these models cover 11 to 110 languages, and only a few Ethiopian languages (2 to 4 languages) are represented due to the lack of large monolingual corpora on the web. Ethiopian languages lack common benchmark datasets for various downstream NLP tasks to evaluate and use for different NLP-related research.  

Our study introduces EthioLLM, a multilingual large language model that accommodates five Ethiopian languages and English. Of these, three languages (Amharic, Ge'ez, and Tigrinya) employ the distinctive Ge'ez writing script, while the remaining two use the Latin script. EthioLLM is developed through the utilization of both XLMR and mT5 architectures in their large, base, and small variants. This multilingual language model is specifically engineered to offer enhanced support for Ethiopian languages by taking into account their diverse scripts and the prevalence of popular languages within the region. 

\section{ EthioLLM}
\subsection{Training Data and Languages}
Even though training LMs requires a large number of datasets \cite{ogueji2021small}, the works by \citet{alabi2022adapting, dossou2022afrolm} showed the possibility of training LMs for languages with a limited amount of data. We followed the same strategy to train EthioLLM as the first step towards developing language models for low-resource Ethiopian languages by collecting available monolingual datasets from different sources for five Ethiopian languages. We collected data from local news media (Fana TV\footnote{\url{https://www.fanabc.com/}}, EBC\footnote{\url{https://www.ebc.et/}}, BBC news\footnote{\url{https://www.bbc.com/}} and Walta\footnote{\url{https://waltainfo.com/}}), the Bible, social media (Facebook and Twitter(X)), and educational textbooks. 


We focused on training our language models with clean data and conducted further pre-processing and cleaning. We also worked on verifying that downstream task training datasets won't end up in the language model training data. Table \ref{tab:languages} shows the selected languages and monolingual dataset used for LMs training.

\begin{table*}[!ht]
\begin{center}
\scalebox{0.7}{
\begin{tabular}{llllrr|p{30mm}cr}
\textbf{Language } &\textbf{script}
& \textbf{Family/branch}& \textbf{\# Speakers} & \textbf{Explored } & \textbf{Data Source}& \textbf{\# Token (M)} &\textbf{\# Sentences} \\
\midrule
Amharic (amh) & Ge'ez & Afro-Asiatic / Ethio-Semitic     & 57M  & yes & $\star, \dagger, \ast$ & 153,509,645 & 9,365,829 \\
English (eng) & Latin & Indo-European / Germanic    & 1268M & yes  & $\star, \ast$  & 76,587,128 & 2,275,996 \\
Afaan Oromo (orm) & Latin & Afro-Asiatic / Cushitic   & 37M & yes  & $\dagger,\star, \ast$  & 22,448,422 & 1,040,175\\
Ge'ez (gez) & Ge'ez & Afro-Asiatic / Ethio-Semitic   & UNK & no  & $\dagger$ & 1,086,578 & 95,899 \\
Somali (som)& Latin & Afro-Asiatic / Cushitic  & 22.3M & no  & $\star, \ast$ & 17,589,974 & 558,161  \\
Tigrinya (tir) & Ge'ez  & Afro-Asiatic / Ethio-Semitic   & 9M & yes  & $\dagger, \star, \ast$  & 28,290,680 & 1,344,586 \\
\bottomrule
\end{tabular}
}
\caption{\textbf{Language model pre-training corpus}: including language family, number of  L1 \& L2 speakers \cite{Eberhard2023}, and number of tokens and sentences for each language. Data source symbols are : $\ast$ = news,  $\star$ = Social media, and $\dagger$ = Spiritual (bible).   
}
\label{tab:languages}
\end{center}
\end{table*}

\subsection{Models}
\subsubsection{Encoder-only models} \label{enc-model}
We trained three multilingual encoder-only models (small, base, and large) with three different parameter configurations. Our encoder-only models used the same parameter setup as AfroXLMR \cite{alabi2022adapting} for all the models. We trained two new tokenizers, one with a 70K vocabulary size and the other one with a 250K vocabulary size. We used a tokenizer with a 70K vocabulary size to train EthioLLM-small and the other one for EthioLLM-base and EthioLLM-large. For EthioLLM-base and EthioLLM-large, the vocabulary sizes are adopted from \citet{alabi2022adapting}, but we wanted to experiment with a smaller vocabulary size for the smaller models to reduce the model size in addition to other hyperparameters.

We adopted the language adaptive fine-tuning (LAFT) strategy proposed by \citet{alabi2022adapting,chi-etal-2021-infoxlm,wang2023nlnde} to train encoder-only models. We tested encoder-only models with different tokenizer sizes starting from 70k-250k for all model variants, but we selected 70K for small and 250K for base and large models based on our initial evaluation in MasakhaneNEWS \cite{adelani2023masakhanews} and MaskahneNER \cite{adelani2021masakhaner} tasks. We also experimented with training from XLMR \cite{conneau2019unsupervised} and AfroXLMR \cite{alabi2022adapting} model checkpoints. Based on our initial task evaluation using similar datasets used in tokenizer evaluation, XLMR \cite{conneau2019unsupervised} model outperformed models trained from AfroXLMR \cite{alabi2022adapting}. 
\subsubsection{Encoder-Decoder models}

To train encoder-decoder models, we adopted the work done by \citet{jude-ogundepo-etal-2022-afriteva}. After sampling from each language, we created 40k vocab size tokenizers for the mt5 small variant model following the Afriteva-small configuration. 

We experimented with different model starting points and observed initializing models from \citet{xue2020mt5} gives better results. Our models are trained for a million steps, and we experimented with different task-specific parameters for different tasks. Our initial assumption that the African-centric models could help if they were used as a starting point did not result in interesting output. We also learned that longer training steps and data cleaning help to get better performance on the small sequence models.


\section{Downstream Tasks and Datasets}
To evaluate our models in diverse downstream tasks, we selected news classification, machine translation, hate speech detection, named entity recognition, part of speech tagging, sentiment analysis, and question analysis tasks. We also created new benchmark datasets for Ethiopian languages (refer to Section \ref{benchmarkdataset}).

\subsection{News Classification}\label{sec:nc}
News classification is one of the text classification problems in NLP, in which news articles are categorized into different classes such as Business, Entertainment, Sports, and others \cite{adelani2023masakhanews}. To address this problem, datasets in four languages (Amharic, Oromo, Tigrinya, and Somali) were collected from publicly available sources. The MasakhaNEWS dataset \cite{adelani2023masakhanews} includes a total of 4,512 news articles categorized into seven different classes. Additionally, a new benchmark dataset was gathered. Specifically, for the Amharic language, 24,265 news articles were obtained from \citet{azime2021}, and 1,875 news articles were sourced from  MasakhaNEWS, resulting in a total of 26,140 articles. Similarly, for the Tigrinya language, 2,397 news articles were obtained from the work of \citet{yohannes2022scheme} and 1,356 news articles were sourced from MaskhaNEWS, resulting in a total of 3,753 articles.

\subsection{Machine Translation (MT)}\label{sec:mt}
MT is a widely used NLP application that automatically translates one language to another to facilitate communication between people who speak different languages \cite{forcada2017making}. 
Many machine translation works \cite{biadgligne2021parallel,gezmu2021extended,teshome2015phoneme,abate2018parallel,teshome2012preliminary,ashengo2021context,ambaye2000} use different statistical machine translation approaches. The work by \citet{belay2022effect} is done by fine-tuning an available multilingual pre-trained model (M2M100 418M) from \citet{nllb2022}. Most of the works use traditional approaches and cover two parallel languages, except for the work of \citet{abate2018parallel}, which covers English and five Ethiopian languages (Amharic, Tigrinya, Afan-Oromo, Wolaytta, and Ge’ez). We combined available MT datasets from the works of \citet{biadgligne2021parallel,abate2018parallel,gezmu2021extended,belay2022effect} and HornMT online repository\footnote{\url{https://github.com/asmelashteka/HornMT}} into one new benchmark dataset.
We present the statistics of the new MT benchmark dataset in Table \ref{tab:newdata}.

\subsection{Hate Speech} \label{sec:hate}
Detecting hate speech plays a crucial role in content moderation by identifying and screening out harmful or offensive language from online platforms, thereby fostering a safer online environment \cite{davidson2017automated,mathew2021hatexplain}. Detecting hate speech in low-resource languages is challenging due to sparse data, linguistic diversity, and complex cultural nuances, making it difficult to develop accurate and contextually aware models \cite{ayelechallenges,ousidhoum-etal-2019-multilingual,muhie2019analysis}.

Based on prior research on Amharic hate speech, we compiled a new benchmark dataset for Amharic comprising approximately 52K data entries sourced from various studies, including 5.3k from \citet{ayele20225js}, 30k from the research by \citet{Tesfaye2020}, 15k from the investigation conducted by \citet{ayelexploringhate2023}. Moreover, for the Afaan Oromo language, we utilized a dataset of 12.8k entries from \citet{ababu-woldeyohannis-2022-afaan}.

\subsection{Sentiment Analysis}\label{sec:sa}
Sentiment analysis constitutes a prominent domain in the field of Natural Language Processing, focusing on the automated detection of emotions or opinions expressed in digital content, including social media posts, blog articles, and reviews. This discipline leverages computational techniques to discern and classify the sentiments or viewpoints encoded in textual data sourced from the internet \cite{agarwal2011sentiment, taboada2011lexicon, yimam-etal-2020-exploring}.


The AfriSenti dataset, as meticulously curated by \citet{muhammad2023AfriSenti}, is designed with a specific focus on African languages. In our research, we harnessed a total of 9,480 Amharic samples and 55,774 samples of Tigrinya. Out of the 55,774 Tigrinya samples, 2,398 were obtained from AfriSenti \cite{muhammad2023AfriSenti}, while the remaining 53,374 samples were sourced from the work of \citet{tela2020transferring}. By amalgamating these two datasets, we created EthioNER dataset as a benchmark dataset for Tigrinya, as elaborated in Section \ref{benchmarkdataset}.


\subsection{Named Entity Recognition (NER)}\label{sec:ner}
Named Entity Recognition (NER) is a fundamental NLP task that involves the identification and classification of predefined information entities within text, which can include proper names, numerical expressions, and temporal references. In our work, we've developed a novel benchmark dataset by amalgamating existing publicly available NER datasets for Amharic, originating from the research of \citet{gamback2017named} and \citet{jibril2023anec}. These two datasets differ in terms of entity classes: \citet{gamback2017named} is annotated with six classes (PER, LOC, ORG, TIME, TTL, and O-other), while \citet{jibril2023anec} features four classes (PER, LOC, ORG, and O-other). To harmonize the classes, we excluded the TIME and TTL categories from \citet{gamback2017named}. Consequently, the new Amharic NER benchmark dataset comprises 292,367 tokens, categorized into four distinct classes. Furthermore, we have created a separate test dataset for the Ge'ez language to evaluate the zero-shot performance of our language models. 

\subsection{Part-of-Speech (POS) Tagging}\label{sec:pos}
POS tagging stands as one of the sequence labeling tasks within the realm of NLP, where each word (token) in a given sentence is assigned a part of speech tag or another philological class \cite{keiper-etal-2016-improving}. To assess the POS tagging capabilities of our models, we employed a publicly available Amharic POS tagging dataset comprising 33,940 sentences (440,941 words) from the research of \citet{gashaw2020machine} and data from the Habit project\footnote{\url{https://habit-project.eu/wiki}} for Amharic, Tigrinya, Oromo, and Somali, with the Habit project data yet to be evaluated by researchers. We merged two Amharic datasets to create a novel benchmark dataset. Additionally, we curated a new Ge'ez POS tagging test dataset to evaluate the zero-shot performance of our models. The statistics of this new benchmark dataset are presented in Table \ref{tab:newdata}.

\subsection{New Benchmark Dataset} \label{benchmarkdataset}
We have amalgamated similar yet independently available datasets into a unified resource, thus creating the \textbf{EthioBenchmark} dataset, tailored for a range of downstream NLP tasks in various Ethiopian languages. While previous research efforts have predominantly focused on individual Ethiopian languages, there remains a dearth of comprehensive downstream task datasets spanning multiple languages of Ethiopia, thereby impeding the progress of future research \cite{tonja2023natural}. The \textbf{EthioBenchmark} dataset has been developed to address this gap and facilitate forthcoming research endeavors in Ethiopian languages.

Henceforth, we will collectively refer to these new benchmark datasets as \textbf{EthioBenchmark}, and designate them as \textit{EthioMMT}, \textit{EthioPOS}, \textit{EthioNEWS}, \textit{EthioHate}, \textit{EthioSenti}, and \textit{EthioNER} for machine translation, POS tagging, news classification, hate speech detection, sentiment analysis, and named entity recognition, respectively. By creating this comprehensive benchmark dataset encompassing multiple Ethiopian languages, we aim to provide a foundation for generating new experimental results that can fuel future analyses in this domain.

For Tigrinya, we have amalgamated existing datasets for machine translation, POS tagging, hate speech detection, and sentiment analysis, thus creating an extensive benchmark dataset. We conducted evaluations using our models to establish baseline results. Comprehensive details regarding \textbf{EthioBenchmark} dataset, encompassing its sources, revised data splitting ratios, and pertinent statistical information, can be found in Table \ref{tab:newdata}. Additionally, we curated new test dataset for Ge'ez by translating sentences from the Amharic NER and POS tagging test sets, resulting in Ge'ez test datasets comprising 1,374 and 1,022 samples for NER and POS tagging, respectively.

\begin{table*}[!ht]
\tiny
\begin{center}
\footnotesize
\begin{tabular}{lllrrrrr}
\toprule
\textbf{NLP Task}  & \textbf{\# Source} & \textbf{Section} &\textbf{amh} & \textbf{orm} &\textbf{som} &\textbf{tir} &\textbf{gez}\\
\midrule
EthioMT & 5 & \ref{sec:mt} & 1,286,902  & 15,484 & 78,426 & 78,426 & 14,720\\
EthioPOS & 2 & \ref{sec:pos} & 22.3M  & 5.3M & 82.4M & 2.7M & 1,022 \\
EthioNEWS & 2 & \ref{sec:nc} & 26,140  & 1,615 & 1,463 & 3,753 & -- \\
EthioSenti & 2 &\ref{sec:sa} & --  & -- & -- & 55,772& --\\
EthioNER & 2 & \ref{sec:ner} & 296,247 & -- & -- & --& 1,374\\
\bottomrule
    \end{tabular}
\caption{\textbf{EthioBenchmark} datasets statistics for each downstream NLP task and language. Under each language category, "-" indicates that we did not compile a new benchmark dataset for that language/task.}
\label{tab:newdata}
  \end{center}
\end{table*}
\section{Results}
We compare the performance of our model against SOTA models that include Ethiopian languages in various downstream tasks using publicly available datasets and newly curated benchmark datasets. 


\subsection{News Classification}
Table \ref{tab:masakhanews_baselines} summarizes different models evaluated on the MasakhaNEWS \cite{adelani2023masakhanews} dataset, using a weighted F1-score as the performance measure. These models are divided into several categories: general multilingual models, Afro-centric models, our encoder-only models, Afro-centric seq2seq models, and our seq2seq models.\\
When comparing the general multilingual models (XLM-R) to the Afro-centric models (AfroXLMR-large and AfroLM), it is clear that the Afro-centric models consistently outperform the general multilingual models for all four languages. AfroXLMR-large achieves higher scores than AfroLM, indicating superior overall performance.

Our encoder-only models (EthioLLM-small, EthioLLM-base, and EthioLLM-large) demonstrate competitive performance compared to the Afro-centric models. In most languages, EthioLLM-small outperforms AfroLM, taking into account parameter size differences. Additionally, EthioLLM-base showed better performance for Amharic and Afan Oromo languages but showed lower performance for Somali and Tigrinya compared to AfroLM.\\
Seq2seq models (AfriTeVa-base and AfriMT5-base) performed less than all encoder-only models across all languages. Our seq2seq model (EthioMT5-small) achieves competitive results compared to the Afro-centric seq2seq models. EthioMT5-small outperforms AfriTeVa-base in all languages and outperforms AfriMT5-base in Amharic, Afaan Oromo, and Tigrinya languages. \\
Overall, our encoder-only models demonstrate competitive performance on the MasakhaNEWS dataset. For seq2seq models, our model outperformed the AfriTeVa-base in all tasks and showed comparative performance with the AfriMT5-base.

\begin{table}[!ht]
\begin{center}
\footnotesize
\small
\begin{tabular}{lllll}
\toprule
\textbf{Model(\#Pram)} & \textbf{amh}  & \textbf{orm} &  \textbf{som}  & \textbf{tir}\\
\midrule
\multicolumn{5}{l}{\textit{SOTA encoder-only models \cite{adelani2023masakhanews}}} \\
XLM-R(550M) &93.1  & 88.4  &76.1 &62.7 \\
AfroXLMR-l(550M) &94.4  &92.1  & 86.9 &89.5 \\
AfroLM (264M) & 90.3 &83.5  &72.0  &83.5 \\ 
\midrule
\multicolumn{5}{l}{\textbf{Our encoder only models}} \\
\rowcolor{Color}
EthioLLM-s (139M) &92.55  &80.84  &64.01 &82.22 \\
\rowcolor{Color}
EthioLLM-b(278M) &91.50  &84.53  &64.78 &76.70 \\  
\rowcolor{Color}
EthioLLM-l(550M) &94.18  &90.89  &77.92 &84.58 \\ 
\midrule
\multicolumn{5}{l}{\textit{SOTA seq2seq models \cite{adelani2023masakhanews}}}\\
AfriTeVa-b(229M) &87.0  &82.9  &58.0 &55.2 \\  
AfriMT5-b(580M) &90.2  &83.9  &77.8 & 80.8   \\
\midrule
\multicolumn{5}{l}{\textbf{Our seq2seq model}  } \\
\rowcolor{Color}
EthioMT5-s (85M)   &90.04  &85.96  &72.44 &82.23 \\
\bottomrule
    \end{tabular}
\caption{\textbf{Baseline results on \masakhanews}.  Evaluation is based on a weighted F1-score. We compared our models with general and Afro-centric models. s = small, b = base, and l = large.}
\label{tab:masakhanews_baselines}
  \end{center}
\end{table}

Table \ref{tab:news_baselines} presents the performance of our models in the new benchmark dataset for Amharic and Tigrinya languages.  As we can see from the table, EthioLLM-large outperformed the other models for Amharic. However, it is important to note that having the highest number of parameters, as seen in EthioLLM-large, does not always guarantee the highest accuracy for both languages. Tigrinya EthioLLM-small outperformed others.

\begin{table}[!h]
\begin{center}
\footnotesize
\small
\begin{tabular}{lrr}
\toprule
\textbf{Model(\#Pram)} & \textbf{amh}  & \textbf{tir}\\
\midrule
\multicolumn{3}{l}{\textbf{Our encoder only models}} \\
\rowcolor{Color}
EthioLLM-s (139M)) &86.53  &83.84   \\
\rowcolor{Color}
EthioLLM-b(278M) & 87.28 & 79.51 \\  
\rowcolor{Color}
EthioLLM-l (550M) &88.94  &83.31    \\ 
\midrule
    \end{tabular}
\caption{\textbf{Baseline results on EthioNEWS dataset}.  Evaluation is based on a weighted F1-score. We only evaluated with our multilingual models. s = small, b = base, and l = large.}
\label{tab:news_baselines}
  \end{center}
\end{table}

\subsection{Sentiment Analysis}
Table \ref{tab:sent_baselines} summarizes the evaluation results for general, Afro-centric, and our models. For Amharic and Tigrinya we utilized the AfriSenti \citet{muhammad2023AfriSenti} dataset.
Additionally in Table \ref{tab:ethiosent_baselines} for Tigrinya, we conducted evaluations across all our models using the new \textbf{EthioSenti} benchmark dataset.

For Amharic, XLMR-large, AfroLM-large, and EthioLLM-large exhibited similar results, achieving an F1 score of approximately 61\%, while EthioLLM-base outperformed AfroLM with an F1 score of 58\%. Among the sequence-to-sequence models, Amharic results show the EthioMT5-small model outperformed the AfriMT5-base, achieving an F1 score of 51.6\%.  For Tigrinya zero shot task AfriMT5-base outperformed EthioMT5-small with an F1 score of 36.9\%.

For \textbf{EthioSenti} Tigrinya results, EthioLM-small outperformed all encoder-only models, attaining an F1 score of 91\%, while EthioLLM-base and large demonstrated comparable results.  
\begin{table}[h!]
\begin{center}
\centering
\begin{tabular}{lrr}
\toprule
\textbf{Model(\#Pram)} & \textbf{amh}  & \textbf{tir$\ast$}\\
\midrule
\multicolumn{2}{l}{\textit{SOTA encoder-only models}} \\
\multicolumn{2}{l}{\cite{muhammad2023AfriSenti}} \\
XLMR-l (550M) &61.8 & -- \\
AfroXLMR-l (550M) &61.6 &62.6 \\
\midrule
\multicolumn{2}{l}{\textbf{Our encoder only models}} \\
\rowcolor{Color}
EthioLLM-s (139M) &56.38  &38.05\\		
\rowcolor{Color}
EthioLLM-b (278M) &58.12  &35.74 \\
\rowcolor{Color}
EthioLLM-l (550M) &61.21  &41.52  \\
\midrule
\multicolumn{2}{l}{\textit{Afro-centric seq2seq LM}} \\
AfriMT5-b  (580M) & 49.4  & 36.9 \\
\midrule
\multicolumn{2}{l}{\textbf{Our seq2seq model}} \\
\rowcolor{Color}
EthioMT5-s  (85M) & 51.6 &   29.5 \\
\bottomrule
    \end{tabular}
\caption{\textbf{Sentiment analysis baseline results on AfriSenti corpus}.  Evaluation is based on a weighted F1-score. s = small, b = base, and l = large. $\ast$= zero-shot performance using Amharic as source language}
\label{tab:sent_baselines}
  \end{center}
\end{table}

\begin{table}[h!]
\begin{center}
\centering
\begin{tabular}{lrr}
\toprule
\textbf{Model(\#Pram)} &  \textbf{tir}\\
\midrule
\multicolumn{2}{l}{\textbf{Our encoder only models}} \\
\rowcolor{Color}
EthioLLM-s (139M) & 91.09\\		
\rowcolor{Color}
EthioLLM-b (278M) & 89.24 \\
\rowcolor{Color}
EthioLLM-l (550M) & 88.86  \\

\bottomrule
    \end{tabular}
\caption{\textbf{Sentiment analysis baseline results on EthioSenti corpus}.  Evaluation is based on a weighted F1-score. s = small, b = base, and l = large.}
\label{tab:ethiosent_baselines}
  \end{center}
\end{table}

\subsection{Hate speech}
To assess our model's performance, we evaluated and compared against the SOTA model. We employed two language datasets provided by \citet{ayele20225js}, \citet{Tesfaye2020}, \citet{ayelexploringhate2023}, \citet{abebaw2022design} and \citet{ababu-woldeyohannis-2022-afaan} for our evaluation. We tested Afro-centric models such as AfroXLMR-large, Afro LM, and the general multilingual model XLM-R for the Amharic and Afan Oromo languages.  


Table \ref{tab:hate_baselines} summarizes the hate speech results for Amharic and Afaan Oromo. As shown in the table, EthioLLM-large outperformed other models for both languages with an F1-score of 73\% and 87\%, respectively, whereas EthioLLM-small and base showed comparable results.

\begin{table}[h!]
\begin{center}
\footnotesize
\begin{tabular}{lrr}
\toprule
\textbf{Model(\#Pram)} & \textbf{amh}  & \textbf{orm}\\
\midrule
\multicolumn{3}{l}{\textit{General multilingual models}} \\
XLM-R (550M) &31.06 &82.89   \\
\midrule
\multicolumn{3}{l}{\textit{Afro-centric models}} \\
AfroXLMR-l (550M) &67.73  & 83.87 \\
AfroLM (264M) &61.69   & 81.40 \\ 
\midrule
\multicolumn{3}{l}{\textbf{Our encoder only models}} \\
\rowcolor{Color}
EthioLLM-s (139M) & 60.90 &84.68 \\
\rowcolor{Color}
EthioLLM-b(278M) & 64.81  &83.24 \\  
\rowcolor{Color}
EthioLLM-l (550M) & \textbf{73.54}  & \textbf{87.28}  \\ 
\midrule
\end{tabular}
\caption{\textbf{Baseline results on EthioHate dataset}.  Evaluation is based on a weighted F1-score. We compared our multilingual models with other models. s = small, b = base, and l = large. }
\label{tab:hate_baselines}
\end{center}
\end{table}

\subsection{Named Entity Recognition (NER)}
We evaluated our models in the NER task using the MasakhaNER dataset \cite{adelani2021masakhaner}, which is a publicly available, high-quality dataset for NER in ten African languages, including only Amharic from Ethiopian languages. For Ge'ez language, we prepared a new NER test set.  Table \ref{tab:MaskhaneNER_baselines}  shows the performance of our models in the Amharic NER task with SOTA models comparison. As we can see from the result, EthiLM-large outperformed all other models with an F1-score of 79\%, while EthioLLM-small and base showed comparable results.

\begin{table}[!h]
\begin{center}
\footnotesize
\small
\begin{tabular}{lr}
\toprule
\textbf{Model(\#Pram)} & \textbf{amh}  \\
\midrule
\multicolumn{2}{l}{\textit{SOTA models} } \\
XLM-R \cite{alabi2022adapting} (550M) &76.18   \\
AfroXLMR-l (550M) \cite{alabi2022adapting} &78.0     \\
AfroLM (264M)\cite{dossou2022afrolm}  &73.84     \\ 
\midrule
\multicolumn{2}{l}{\textbf{Our encoder only models}} \\
\rowcolor{Color}
EthioLLM-s (139M)) & 68.99  \\
\rowcolor{Color}
EthioLLM-b(278M) & 69.9  \\  
\rowcolor{Color}
EthioLLM-l (550M) & \textbf{79.42}    \\ 
\midrule
    \end{tabular}
\caption{\textbf{Baseline results on our MaskhaneNER dataset}.  Evaluation is based on a weighted F1-score. We compared our multilingual models with others. s = small, b = base, and l = large.}
\label{tab:MaskhaneNER_baselines}
  \end{center}
\end{table}

In Table \ref{tab:ner_baselines}, we evaluated our models in EthioNER datasets. We used Amharic as the source language to evaluate Ge'ez's zero-shot performance. For Amharic, EthioLLM-large outperformed base and small models with an F1-score of 78\%, while EthioLLM-small and base have shown comparable results. All models have shown a promising result for Ge'ez zero-shot task, while EthioLLM-large outperformed the rest with an F1-score of 74\%.
\begin{table}[!h]
\begin{center}
\footnotesize
\small
\begin{tabular}{lrr}
\toprule
\textbf{Model(\#Pram)} & \textbf{amh}  & \textbf{gez*}\\
\midrule
\multicolumn{2}{l}{\textbf{Our encoder only models}} \\
\rowcolor{Color}
EthioLLM-s (139M)) & 71.83 &73.67 \\
\rowcolor{Color}
EthioLLM-b(278M) & 73.06  & 73.79 \\  
\rowcolor{Color}
EthioLLM-l (550M) & \textbf{78.02}  & \textbf{74.84}  \\ 
\midrule
    \end{tabular}
\caption{\textbf{Baseline results on EthioNER dataset}.  Evaluation is based on a weighted F1-score. * shows the zero-shot performance using Amharic as source language. s = small, b = base, and l = large.}
\label{tab:ner_baselines}
  \end{center}
\end{table}

\subsection{Part of Speech Tag (POS)}
Table \ref{tab:pos_baselines} shows the results of our models. We evaluated the model on our benchmark dataset. Our EthioLLM-large model archives 90.36\%, 99.98\%, and 79.67\% on amh, orm, and tir tasks, respectively.

\begin{table}[!h]
\begin{center}
\footnotesize
\small
\begin{tabular}{lrrrr}
\toprule
\textbf{Model(\#Pram)} & \textbf{amh}  & \textbf{orm} & \textbf{tir} & \textbf{gez*}\\
\multicolumn{4}{l}{\textbf{Our encoder only models}} \\
\rowcolor{Color}
EthioLLM-s (139M) &86.86  &99.95 &78.33&35.84  \\
\rowcolor{Color}
EthioLLM-b(278M) &85.09   & 99.95 &71.93& 34.84  \\  
\rowcolor{Color}
EthioLLM-l (550M) &90.36  &99.98 & 79.67 &37.63   \\ 
\midrule
    \end{tabular}
\caption{\textbf{Baseline results on EthiPOS tag dataset}.  Evaluation is based on a weighted F1-score. * shows the zero-shot performance using Amharic as a source language. s = small, b = base, and l = large.}
\label{tab:pos_baselines}
  \end{center}
\end{table}

\subsection{Machine Translation}

\begin{table}[h!]
\begin{center}
\footnotesize
\small
\begin{tabular}{lrrrrrr}
\toprule
\textbf{Source} & \textbf{amh}  & \textbf{eng} & \textbf{orm} &  \textbf{som}  & \textbf{tir} & \textbf{gez}\\
\midrule
\multicolumn{5}{l}{\textit{EthioMT5-S 85M }} \\
amh & - & 17.0 & 0.84 & 0.88 & 0.84 & 5.30 \\
eng & 5.45 & - & 1.30 & 2.60 & 0.70 & 0.70 \\
\midrule
\multicolumn{5}{l}{\textit{M2M100 418M}} \\
amh & - & 37.60 & $\ast$ & 2.90 & 2.86 & $\ast$ \\
eng & 13.70 & - & $\ast$ & 9.60 & 9.60 & $\ast$ \\

\bottomrule
    \end{tabular}
\caption{\textbf{Baseline sacreBleu results of EthioMT} on Flores-200 \cite{nllb2022} for languages except for Ge'ez. $\ast$ = languages not covered in M2M100. Results of M2M100 are from \cite{nllb2022} paper for \textit{eng-xx} and \textit{xx-eng} model, and we fine-tuned for the others. }
\label{tab:mt_baselines}
  \end{center}
\end{table}

Table \ref{tab:mt_baselines} presents baseline results for EthioMMT dataset. We utilized the Flores-200 dataset \cite{nllb2022} for evaluation across all language pairs except for Ge'ez. For Ge'ez, we created our test split and subsequently reported the results on this custom test split. To compare the performance of our models, we used \citet{nllb2022} results for the languages mentioned in the paper and finetuned for the rest. 

\citet{nllb2022} is the state-of-the-art in MT, but we showed the closeness we can achieve to the model with a smaller MT5 model. This model can be a good experimental platform for MT tasks with fewer trainable parameters. This lower score shown in machine translation by our MT5 models is also observed in models like Afriteva and AfriMT5. Our model also covers two previously uncovered languages in \citet{nllb2022}, which we found beneficial in Ethiopian languages.

\section{Discussion}
We compared our models with current SOTA models that include Ethiopian languages. Our models show comparable results with SOTA models. From our models, the EthioLLM-large model shows comparable results in news classification and sentiment analysis tasks and outperforms the existing SOTA model in named entity recognition and hate speech tasks. EthioLLM-small with a parameter size of 139M showed comparable results with AfroLM \cite{dossou2022afrolm} and outperformed XLM-R \cite{conneau2019unsupervised} in sentiment analysis and hate speech detection.     

We showed that our EthioMT5-small model performs better or is on par with the other base models on the classification tasks. This can be attributed to the longer training and data cleaning we did to train our language model. The same explanation doesn't work for tasks like machine translation, where our model fails short compared to m2m100 models. This is understandable given the smaller size of the model, but for machine translation tasks, the best approach would be to fine-tune m2m100 models directly. 

Our models exhibit promising results in zero-shot evaluation for Ge'ez, suggesting that they may also perform well for low-resource languages incorporated during language pre-training. We released\footnote{\url{https://github.com/EthioNLP/EthioLLM}}  the EthioLLM models, EthioBenchmark dataset, and our top-performing task-specific models as open-source resources, aiming to encourage further research in Ethiopian languages.


\section{Conclusion and Future Work}
In this paper, we presented EthioLLM, the first attempt to train multilingual language models for five Ethiopian languages and English. We tested our EthiLLM with the available benchmark datasets like MasakhaNEWS, MasakhaNER, and AfriSenti. We also created \textbf{EthioBenchmark} dataset for various downstream tasks for five Ethiopian languages by combining the available corpus. Additionally, we created a new task for Ge'ez. We included a minimum of two downstream tasks for each language in the language models. Our models have outperformed some and demonstrated comparable performance with respect to the current SOTA models in different cases. 

As shown in the results section, our sequence-to-sequence models were tested with only a machine translation sequence-to-sequence task. In addition to the tasks we tried, we plan to train both the base and large versions of these models and introduce several other sequence-to-sequence tasks apart from machine translation.


\section*{Limitations}
In this work, we presented models and downstream task evaluation for five Ethiopian languages with a publicly available evaluation dataset and created a new benchmark dataset as one of the contributions for the languages left behind by current technology. 
Despite our efforts, a significant gap exists in the downstream task creation, spanning multiple languages. The primary challenge lies in developing a diverse set of tasks that encompass all languages within the language model. Another challenge we encounter is acquiring a sufficient amount of data for language model training. Due to the scarcity of corpora. There are more than 85 languages in Ethiopia but we only covered 5 of them in this study because of the scarcity of corpus.


\nocite{*}
\section{Bibliographical References}\label{sec:reference}

\bibliographystyle{lrec-coling2024-natbib}
\bibliography{lrec-coling2024-example}

\end{document}